\documentclass[lettersize, journal]{IEEEtran}

\usepackage{amsmath}
\usepackage{amssymb}
\usepackage{amsfonts}
\usepackage{cite}
\usepackage{subfigure}
\usepackage[pdftex]{graphicx}
\usepackage[acronym, nonumberlist]{glossaries}
\usepackage{url}
\usepackage{threeparttable}
\usepackage[commandnameprefix=ifneeded]{changes}
\usepackage{comment}
\usepackage{physics}
\usepackage{cleveref}
\usepackage[OT1]{fontenc} 

\definechangesauthor[name={andrey}, color=blue]{AP}

\newacronym{lstm}{LSTM}{Long-Short Term Memory}
\newacronym{gru}{GRU}{Gated Recurrent Unit}
\newacronym{rnn}{RNN}{Recurrent Neural Network}
\newacronym{sac}{SAC}{Soft Actor-Critic}
\newacronym{vae}{VAE}{Variational Autoencoder}
\newacronym{bev}{BEV}{Bird-Eye View}
\newacronym{cnn}{CNN}{Convolutional Neural Network}
\newacronym{rl}{RL}{Reinforcement Learning}
\newacronym{ddpg}{DDPG}{Deep Deterministic Policy Gradient}
\newacronym[longplural={Artificial Neural Networks}]{ann}{ANN}{Artificial Neural Network}
\newacronym{mdn}{MDN}{Mixture Density Network}
\newacronym{gan}{GAN}{Generative Adversarial Network}
\newacronym{mlp}{MLP}{Multi-Layer Perceptron}
\newacronym{imu}{IMU}{Inertial Measurement Unit}
\newacronym{dqn}{DQN}{Deep Q-Network}
\newacronym{mse}{MSE}{Mean Squared Error}
\newacronym{msssim}{MS-SSIM}{Multi-Scale Structural Similarity Index Measure}
\newacronym{a2c}{A2C}{Advantage Actor Critic}
\newacronym{ppo}{PPO}{Proximal Policy Optimization}
\newacronym{ae}{AE}{Autoencoder}
\newacronym{dae}{DAE}{Dynamic Autoencoder}
\newacronym{cdae}{CDAE}{Combined Dynamic Autoencoder}
\newacronym{CARNet}{CARNet}{\textbf{C}ombined dyn\textbf{A}mic autoencode\textbf{R} \textbf{Net}work}
\newacronym{KARNet}{KARNet}{\textbf{K}alman Filter \textbf{A}ugmented \textbf{R}ecurrent Neural \textbf{Net}work}
\newacronym{dvae}{DVAE}{Dynamic Variational Autoencoder}
\newacronym{wm}{WM}{World Models}
\newacronym{mdnrnn}{MDN-RNN}{Mixture Density Recurrent Neural Network}
\newacronym{mpc}{MPC}{Model-Predictive Control}
\newacronym{il}{IL}{Imitation Learning}

\newcommand*\Bell{\ensuremath{\boldsymbol\ell}}

\begin{document}

\title{KARNet: Kalman Filter Augmented Recurrent Neural Network for Learning World Models in Autonomous Driving Tasks}

\author{Hemanth Manjunatha$^1$, \emph{Student Member, IEEE}, 
Andrey Pak$^1$, \emph{Student Member, IEEE},\\  
	Dimitar Filev$^2$, \emph{Fellow, IEEE}, 
	Panagiotis Tsiotras$^1$, \emph{Fellow, IEEE} 
	\thanks{$^1$School of Aerospace Engineering and Institute for Robotics and Intelligent Machines, Georgia Institute of Technology, Atlanta, GA 30332-1050.
	Atlanta, GA 30332-1050.}
	\thanks{$^2$Research \& Advanced Engineering, Ford Motor Company,
	Dearborn, MI 48121.}
}

\maketitle

\begin{abstract}
	Autonomous driving has received a great deal of attention in the automotive industry and is often seen as the future of transportation. 
	The development of autonomous driving technology has been greatly accelerated by the growth of end-to-end machine learning techniques that have been successfully used for perception, planning, and control tasks.
	An important aspect of autonomous driving planning is knowing how the environment evolves in the immediate future and taking appropriate actions. 
	An autonomous driving system should effectively use the information collected from the various sensors to form an abstract representation of the world to maintain situational awareness. 
	For this purpose, deep learning models can be used to learn compact latent representations from a stream of incoming data.
	However, most deep learning models are trained end-to-end and do not incorporate in the architecture any prior knowledge (e.g., from physics) of the vehicle. 
	In this direction, many works have explored physics-infused neural network (PINN) architectures to infuse physics models during training. 
	Inspired by this observation, we present a Kalman filter augmented recurrent neural network architecture to learn the latent representation of the traffic flow using front camera images only.
	We demonstrate the efficacy of the proposed model in both imitation and reinforcement learning settings using both simulated and real-world datasets.
	The results show that incorporating an explicit model of the vehicle (states estimated using Kalman filtering) in the end-to-end learning significantly increases performance.
	 
\end{abstract}

\glsreset{ae}

\section{Introduction}

Agents that can learn autonomously while interacting with the world are quickly becoming mainstream due to advances in the machine learning domain \cite{amershi2014power}. 
Particularly, stochastic generative modeling and reinforcement learning frameworks have proved successful in learning strategies for complex tasks, often out-performing humans by learning the structure and statistical regularities found in data collected in the real world \cite{matsuo2022deep, friston2021world}. 
These successful theoretical frameworks support the idea that acquiring internal models of the environment, i.e. World Models (WM) is a natural way to achieve desired interaction of the agent with its surroundings.
Extensive evidence from recent neuroscience/cognitive science research \cite{downing2009predictive, svensson2013dreaming} highlights prediction as one of the core brain functions, even stating that ``prediction, not behavior, is the proof of intelligence'' with the brain continuously combining future expectations with present sensory inputs during decision making.
One can argue that prediction provides an additional advantage from an evolutionary perspective, where the ability to predict the next action is critical for successful survival. 
Thus, constructing explicit models of the environment, or enabling the learning agent to predict its future, has a fundamental appeal for learning-based methods \cite{kaiser2019model}. 
In this direction, we present a physics-infused neural network architecture for learning World Models for autonomous driving tasks. 
By World Model we mean predicting the future frames from consecutive historical frames where a frame is the front view image of the traffic flow~\cite{wang2019memory}.

\gls{rl} and \gls{il} have gained much traction in autonomous driving as a promising avenue to learn an end-to-end policy that directly maps sensor observations to steering and throttle commands. 
However, approaches based on \gls{rl} (or \gls{il})  require many training samples collected from a multitude of sensors of different modalities, including camera images, LiDAR scans, and Inertial-Measurement-Units (IMUs).
These data modalities generate high-dimensional data that are both spatially and temporally correlated. 
In order to effectively use the information collected from the various sensors and develop a world model (an abstract description of the world), this high-dimensional data needs to be reduced to low-dimensional representations~\cite{lesort2018state}.
To this end, \glspl{vae} and \glspl{rnn} have been extensively used to infer low-dimensional \textit{latent variables} from temporally correlated data \cite{ha2018recurrent, lipton2015critical, salehinejad2017recent, plebe2020road}.
Furthermore, modeling temporal dependencies using \gls{rnn} in the data allows to construct a good \textit{prediction model} (a World Model), which 
has been shown to benefit \gls{rl}/\gls{il} scenarios~\cite{werbos1987learning, silver2017predictron}.
However, learning a good prediction model in an end-to-end fashion comes with the cost of substantial training data, training time, not to mention a significant computational burden. 
Moreover, end-to-end training can lead to hallucinations where an agent learns World Models that do not adhere to physical laws~\cite{ha2018recurrent, hallucinations}. 
Hence, there is a need to consider physics-informed or model-based machine learning approaches \cite{karniadakis2021physics}.

Model-based learning methods, on the other hand, incorporate prior knowledge (e.g., from physics) into the neural network architecture~\cite{karniadakis2021physics,cuomo2022scientific}.
The explicit model acts as a high-level abstraction of the observed data and can provide rich information about the process, which might not be possible by directly learning from limited data~\cite{meng2022physics}. 
In addition, model-based methods can significantly improve data efficiency and generalization~\cite{meng2022physics, moerland2023model}. 
Nonetheless, model-based methods introduce bias due to the fact that the hand-crafted model cannot capture the complete temporal characteristics. 
Therefore, a generally accepted belief is that model-free methods are data-hungry but achieve better asymptotic performance. 
On the other hand, model-based algorithms are more data efficient but achieve sub-optimal performance~\cite{pong2018temporal}. 
Hence, there is an imperative need to combine the best of both these approaches.

\begin{figure*}[t!] 
	\centering
	{\includegraphics[width=0.9\linewidth]{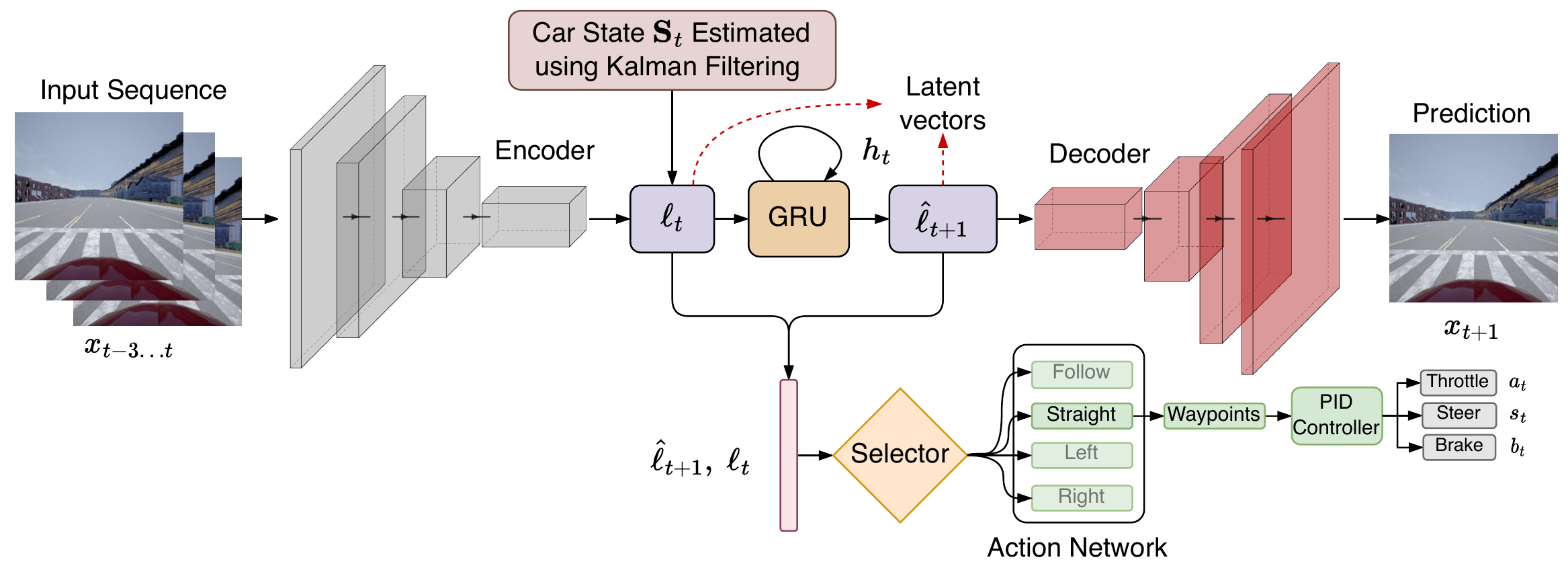}
		\caption{Overview of the \gls{KARNet} architecture with imitation and reinforcement learning. 
			The proposed \gls{KARNet} combines both the encoder and decoder networks with a recurrent neural network to learn present and future latent vectors. 
		These latent vectors are used to learn autonomous driving tasks using reinforcement learning.}
		\label{fig:overview}}
\end{figure*}

Motivated by these observations, we present a combined model for learning  World Models from high-dimensional, temporally correlated data for driving tasks.
The proposed combined architecture is shown in Fig.~\ref{fig:overview}. 
The architecture combines an \gls{ae}, an \gls{rnn} (in the form of \gls{gru}), and a Kalman filter~\cite{kalman1960new} in a single network. 
The \gls{ae} and \gls{rnn} are trained together as a single network.
The \gls{ae} is used to learn the latent representation from the images and the \gls{rnn} is used to predict the next latent variable given the present latent representation.
Many previous works support the rationale that combined training of an inference model (\gls{ae}) and a generative model (\gls{rnn}) can improve the performance.
\cite{girin2020dynamical, salimans2016structured, chung2015recurrent, kingma2019introduction}.  
Throughout this paper, we refer to this combined model as \gls{KARNet}. 
The \gls{KARNet} learns to predict the next frame of the traffic given a history of the previous frames by learning a latent vector from raw data. 
These latent vectors are then used to learn autonomous driving tasks using reinforcement learning.
Learning the latent vectors has a grounding in neuroscience called \textit{modal} symbols~ \cite{kiefer2012conceptual}, which can be thought of as states of the sensorimotor systems. 
These modal symbols can be used to re-enact the perception and action that produced those symbols. 
Along the same lines, the \textit{latent vector space} can be considered a modal symbol extracted from the images and the vehicle's state and is used to learn the driving behavior.

\subsection*{Contributions}

We present the \textbf{K}alman Filter \textbf{A}ugmented \textbf{R}ecurrent Neural \textbf{Net}work (KARNet) that learns current and future latent representations from raw high-dimensional data and, in addition, acts as a future latent vector prediction model.
In contrast to similar end-to-end driving architectures (e.g., \cite{ha2018recurrent}), our approach combines model-based Kalman filtering to integrate vehicle state information with a data-driven neural network architecture. 
The central intuition behind the proposed approach is that although we do not have a concrete mathematical model to predict the traffic flow from the next front camera images, and thus we need to use
data-driven end-to-end methods, on the other hand, we do have an accurate model of the motion of the car itself. 
Hence, we can leverage conventional filtering approaches to estimate the vehicle's state and combine it with an end-to-end learning network to better predict the traffic. 
As we wish to study the efficacy of combining model-based and model-free approaches,
we use the model-free approach as a baseline for comparison and also investigate where the model-based information should be integrated with a model-free framework to improve performance.
Concretely, this study explores two questions, specifically in the context of autonomous driving tasks: 
First, does integrating model-based information with model-free end-to-end learning improve performance? and, second, how to best integrate the model-based information in a model-free architecture?
We validate the efficacy of the proposed integration framework using via two experiments using imitation learning and reinforcement learning.

\section{Related Work}

Knowing how the environment may evolve in the immediate future is crucial for the successful and efficient planning of an agent interacting 
with a complex environment. 
Traditionally, this problem has been approached using \gls{mpc} along with system identification~\cite{swief2019survey}. 
While these classical approaches are robust and well-established  when the dynamical model is known and relatively simple, 
in practice, applying these techniques to more realistic and complex problems, such as autonomous driving, becomes increasingly difficult.

\subsection{Insights from Cognitive Science}

Research in neuroscience and cognitive science emphasizes prediction as a fundamental brain function \cite{downing2009predictive}. 
In fact Llinas~\cite{llinas2002vortex} and Hawkins~\cite{hawkins2004intelligence} argue that prediction is the pinnacle of cerebral functionality --- which is the core of simple and sophisticated intelligence. 
There are many predictive mechanisms studies in neuroscience as summarised by Downing \cite{downing2009predictive}. 
The two important ones are located in the cerebellum and in the dorsal/ventral stream in the brain with completely different neural architectures serving different purposes. 
In our case, the proposed \gls{KARNet} architecture with an \gls{ae} and \gls{rnn} is close to the Convergence-Divergence zone model (CDz) of the dorsal stream \cite{meyer2009convergence}.
However, the CDz architecture combines several encoders progressively, while the proposed \gls{KARNet} architecture has only one encoder and one decoder.
Moreover, learning the predictive model and the control has many parallels in learning via mental imagery \cite{da2020mental}.

\subsection{Sequence Modelling}
This section overviews the current approaches addressing learning, simulating, and predicting dynamic environments.
First, we briefly cover the literature on learning the environment using model-free or end-to-end learning methods. 
Next, we cover the model-based methods, which specifically use Kalman filtering.

\glsreset{rnn} \glspl{rnn} are among the most popular architectures frequently used to model time-series data. 
The idea of using \glspl{rnn} for learning the system dynamics, making future predictions, and simulating previously observed environments, has recently gained popularity~\cite{wahlstrom2015pixels,watter2015embed,patraucean2016spatiotemporal,sun2016learning}.
The capability of encoding/generative models, combined with the predictive capabilities of the recurrent networks, 
provides a promising framework for learning complex environment dynamics. 
In this regard, Girin et al~\cite{girin2020dynamical} provide a broad overview of available methods that model the temporal dependencies within a sequence of data and the corresponding latent vectors. 
This overview covers the recent advancements in learning and predicting environments such as basic video games and frontal-camera autonomous driving scenarios based on combining an autoencoder and a recurrent network.

Generative recurrent neural network architectures have also been used in simulators.
In \cite{chiappa2017recurrent}, the authors propose a recurrent neural network-based simulator that learns action-conditional-based dynamics that can be used as a surrogate environment model while decreasing the computational burden. 
The recurrent simulator was shown to handle different environments and capture long-term interactions. 
The authors also show that state-of-the-art results can be achieved in Atari games and a 3D Maze environment using a latent dynamics model learned from observing both human and AI playing.
The authors also highlighted the limitations of the proposed \gls{rnn} architecture for learning more complex environments (such as 3D car racing and 3D Maze navigation) and identified the existing trade-offs between long and short-term prediction accuracy. 
They also discussed and compared the implications of prediction/action/observation dependent and independent transitions. 

In terms of learning explicit World Models (WM),
Ha and Schmidhuber~\cite{ha2018recurrent} propose an end-to-end trained machine learning approach to learn the world dynamics using \gls{vae} and \gls{rnn} in a reinforcement learning context. 
They showed how spatial and temporal features extracted from the environment in an unsupervised manner could be successfully used to infer the desired control input and how they can yield a policy sufficient for solving typical \gls{rl} benchmarking tasks in a simulated environment. 
The proposed WM architecture in \cite{ha2018recurrent} consists of three parts. 
First, a \gls{vae} is used to learn the latent representation of several simulated environments (e.g., Car Racing and VizDoom);
second, a \gls{mdn}-\gls{rnn} \cite{graves2014generating}  performs state predictions based on the learned latent variable and the corresponding action input; and, finally, 
a simple \gls{mlp} determines the control action from the latent variables. 
The authors achieved state-of-the-art performance by training the model in an actual environment and the simulated latent space, i.e., the ``dream'' generated by sampling the \gls{vae}'s probabilistic component.

In \cite{hafner2019learning} the authors addressed the challenge of learning the latent dynamics of an unknown environment. 
The proposed PlaNet architecture uses image observations to learn a latent dynamics model and then subsequently performs planning and action choices in the latent space. 
The latent dynamics model uses deterministic and stochastic components in a recurrent architecture to improve planning performance.

In a similar study from Nvidia, GameGAN \cite{kim2020learning} proposed a generative model that learns to visually imitate the desired game by observing screenplay and keyboard actions during training. 
Given the gameplay image and the corresponding keyboard input actions, the network learned to imitate that particular game. 
The approach in \cite{kim2020learning} can be divided into three main parts: 
The Dynamics Engine, an action-conditioned \gls{lstm} \cite{hochreiter1997long,chiappa2017recurrent} that learns action-based transitions; the external Memory Module, whose objective is to capture and enforce long-term consistency; and the Rendering Engine that renders the final image not only by decoding the output hidden state but also by decomposing the static and dynamic components of the scene, such as the background environment vs. moving agents. 
The authors achieve state-of-the-art results for some game environments such as Pacman and VizDoom~\cite{kempka2016vizdoom}.

The recent DriveGAN \cite{kim2021drivegan} approach proposes a scalable, fully-differentiable simulator that also learns the latent representation of a given environment. 
The latent representation part of the DriveGAN uses a combination of $\beta$-\gls{vae} \cite{higgins2016beta} for inferring the latent representation and StyleGAN \cite{karras2019style,karras2020analyzing} for generating an image from the latent vector. 
This approach has the advantage over classic \gls{vae} that  also learns
 a similarity metric between the produced images during the training process~\cite{larsen2016autoencoding}.

The combination of end-to-end deep learning models with model-based methods, specifically Kalman-based algorithms, is gaining a lot of attention for learning non-linear or unknown dynamics \cite{krishnan2015deep, zhou2020kfnet, barratt2020fitting, coskun2017long, revach2022kalmannet}. 
These approaches are used to learn the filtering parameters (e.g., noise, gains).
Moreover, a parametric model is assumed in the above strategies, and the parameters are estimated using data. 
However, when the dynamics of the system are complex or cannot be captured in closed form (e.g., visual observations in autonomous driving tasks), one needs to resort to data-driven deep learning models.

In summary, current state-of-the-art end-to-end techniques use two main components to perform accurate predictions. 
First, an unsupervised pre-training step is utilized to learn the latent space of the given environment.
This step typically involves an autoencoder or an autoencoder–\gls{gan} combination. 
Second, a recurrent network, usually in the form of \gls{lstm}, is used to capture the spatiotemporal dependencies in the data and make future predictions based on the previously observed frames and action inputs. %
However, most of these prior works train the autoencoder and recurrent architectures separately, with one of the main reasons being the training speed. Although this training approach is perfectly reasonable and achieves state-of-the-art results, some dynamic relations captured by the recurrent part might not be rich-enough.
Moreover, the above works use end-to-end learning and do not explicitly use model-based methods in their architecture.
On the other hand, the Kalman-algorithm-based approaches focus on learning the non-linear or unknown dynamics of a system from the data. 
In our approach, we use a Kalman filter to model the known vehicles' dynamics and the end-to-end deep learning approach to model the environment.

\section{Approach}
\label{sec:approach}

\begin{figure}[t!] 
	\centering
	\includegraphics[width=\linewidth]{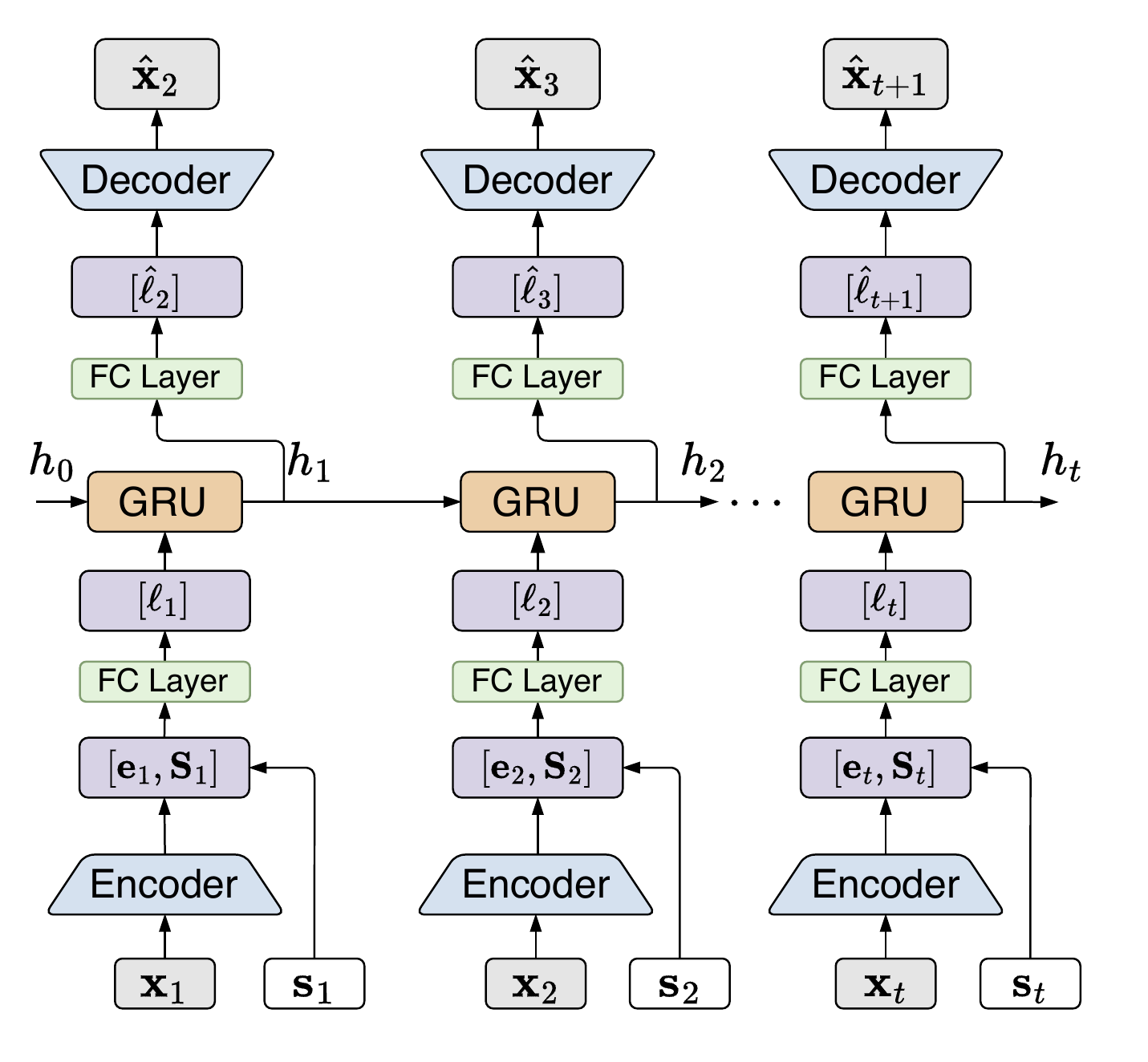}
	\caption{\gls{KARNet} architecture. A sequence of images is first encoded into latent space, fused (concatenated) with corresponding sensor measurements, then propagated through the recurrent network and outputs the estimate of both latent state and sensor measurements at timestep $t$.}
	\label{fig:combined_architecture_all}
\end{figure}

In the following sections, we first examine the architecture of \gls{KARNet} followed by early and late-fusion of  the ego vehicle's state estimated by an extended Kalman filter. 
Second, we discuss the training procedure, data generation, and bench-marking setup. 

\subsection{KARNet Architecture}
\label{sec:cdae_architecture}
The overall architecture of the proposed \gls{KARNet} is shown in Fig.~\ref{fig:combined_architecture_all}. 
The architecture combines \gls{ae}-\gls{rnn} neural networks trained together to make accurate predictions of future traffic image frames using latent representations.
The proposed approach is inspired by the World Models (WM) \cite{ha2018recurrent} architecture.
However, in contrast with WM, where the \gls{vae} and \gls{rnn} do not share any parameters, in our approach, the \gls{ae} and \gls{rnn} networks are trained together. 
The main intuition behind training together the \gls{ae} and \gls{rnn} networks is as follows: 
while the autoencoder itself provides sufficient capabilities for capturing the latent state and reconstructing the subsequent images, 
it may omit temporal dependencies necessary for properly predicting the latent variable. 
In addition to operating purely on latent vectors, the architecture can also incorporate the vehicle state (estimated using conventional filtering methods) with the latent/hidden vectors (see Fig.~\ref{fig:combined_architecture_all}). 
The rationale is that images alone are not sufficient to capture complex environments (e.g., traffic) 
and additional inputs may be needed.
Moreover, autonomous vehicles are already equipped with a wide array of sensors that can estimate the vehicle's state.

\begin{table}[h!]
    \renewcommand{\arraystretch}{1.25}
    \centering
    \caption{Autoencoder structure}
    \label{tab:autoencoder_architecture}
        \begin{tabular}{|c|c|c|c|}
        
        \multicolumn{2}{c}{\textbf{Encoder}}                                                                   & \multicolumn{2}{c}{\textbf{Decoder}}            \\ \hline
        \textbf{Layer}                                                       & \textbf{Output Shape} & \textbf{Layer}                                                         & \textbf{Output Shape}    \\ \hline
        \textbf{Input}                                                       & 1x256x256    & \textbf{Input}                                                         & 1x128     \\ \hline
        \begin{tabular}[c]{@{}l@{}}conv3-2\\ conv3-2\end{tabular}   & 2x128x128    & \begin{tabular}[c]{@{}l@{}}tconv3-64\\ tconv3-64\end{tabular} & 64x4x4    \\ \hline
        \begin{tabular}[c]{@{}l@{}}conv3-4\\ conv3-4\end{tabular}   & 4x64x64      & \begin{tabular}[c]{@{}l@{}}tconv3-32\\ tocnv3-32\end{tabular} & 32x8x8    \\ \hline
        \begin{tabular}[c]{@{}l@{}}conv3-8\\ conv3-8\end{tabular}   & 8x32x32      & \begin{tabular}[c]{@{}l@{}}tconv3-16\\ tconv3-16\end{tabular} & 16x16x16  \\ \hline
        \begin{tabular}[c]{@{}l@{}}conv3-16\\ conv3-16\end{tabular} & 16x16x16     & \begin{tabular}[c]{@{}l@{}}tconv3-8\\ tconv3-8\end{tabular}   & 8x32x32   \\ \hline
        \begin{tabular}[c]{@{}l@{}}conv3-32\\ conv3-32\end{tabular} & 32x8x8       & \begin{tabular}[c]{@{}l@{}}tconv3-4\\ tconv3-4\end{tabular}   & 4x64x64   \\ \hline
        \begin{tabular}[c]{@{}l@{}}conv3-64\\ conv3-64\end{tabular} & 64x4x4       & \begin{tabular}[c]{@{}l@{}}tconv3-2\\ tconv3-2\end{tabular}   & 2x128x128 \\ \hline
        conv3-128                                                   & 128x1x1      & tconv3-1                                                      & 1x256x256 \\ \hline
        \end{tabular}
\end{table}
The architecture of the baseline autoencoder (see Table~\ref{tab:autoencoder_architecture}) follows a simple/transposed convolution architecture with a kernel size of $3\times 3$. All convolution layers are followed with a batch normalization layer and a ReLU activation. With the dimension of the input image being $256\times 256$, the resulting size of the latent variable is $128$. 

To better understand the flow of the processing induced by the proposed architecture (Fig.~\ref{fig:combined_architecture_all}), let us consider a sequence of $t+1$ consecutive frames $\left[\mathbf{x}_1, \ldots,\mathbf{x}_{t+1}\right]$ taken from the frontal camera of a moving vehicle, along with the corresponding sensor data. 
Note that the subscript $1$ refers to the beginning of the current sliding window and not the beginning of the dataset.
Given the first $t$ frames $\left[\mathbf{x}_1, \ldots,\mathbf{x}_{t}\right]$ and the corresponding vehicle's states $\left[\mathbf{s}_1, \ldots, \mathbf{s}_{t}\right]$, our aim is to learn the latent representations $\left[\Bell_1, \ldots, \Bell_{t}\right]$ and predict $\mathbf{x}_{t+1}$.
The vehicle's state is a vector $\mqty[p_x & p_y & v_x & v_y]$ where $p_x$ and $p_y$ are positions measured in meters and $v_x$ and $v_y$ are velocities measured in m/s. 

Referring to Fig.~\ref{fig:combined_architecture_all}, and
for the sake of simplicity, let us consider a single-step prediction at time step $t$, without any vehicle state ($\mathbf{s}_t$)
The image $\mathbf{x}_{t}$ is encoded into the latent variable as $\mathbf{e}_{t}=\mathbf{E}(\mathbf{x}_{t})$, where $\mathbf{E}$ denotes the encoder.  
Also, note that the encoder/decoder can be any architecture or a typical convolutional neural network; hence, we assume a general architecture as follows.
\begin{subequations} \label{eq:encoder-decoder}
	\begin{align} 
		\mathbf{e}_{t}       & = \mathbf{E}(\mathbf{x}_{t}), \\
		\hat{\mathbf{x}}_{t} & = \mathbf{D}(\mathbf{e}_{t}). 
	\end{align}
\end{subequations}
The encoded vector $\Bell_{t} = W_{fc1}\mathbf{e}_t $ is used as the input to the \gls{gru} block, whose output is the latent space vector at the next time step $\hat{\Bell}_{t+1} = W_{fc2}\mathbf{h}_t$, where the $\mathbf{h}_t$ is given by
\begin{subequations}
	\begin{align}
		z_{t}          & = \sigma\left(W_z \left[\mathbf{h_{t-1}}, \, \Bell_{t}\right]\right), \label{eq:gru-a} \\
		r_{t}          & = \sigma\left(W_r  \left[\mathbf{h_{t-1}}, \, \Bell_{t}\right]\right),                 \\
		\tilde{h}_t    & = \tanh\left( W  \left[r_t \odot \mathbf{h_{t-1}}, \, \Bell_{t} \right]\right),        \\
		\mathbf{h_{t}} & = (1-z_t) \odot \mathbf{h_{t-1}} + z_t \odot \tilde{h}_t,                              
	\end{align}
\end{subequations}
where $W_{fc1},\ W_{fc2},\ W,\ W_z,\ W_r$ are  learnable weights, $\sigma$ is a sigmoid function, and $\odot$ is the Hadamard product. 
The predicted latent variable $h_{t}$ is used as the hidden state for the next time step, and the predicted $\hat{\Bell}_{t+1}$ is used for reconstructing the next timestep image, $\hat{\mathbf{x}}_{t+1} = \mathbf{D}(\hat{\Bell}_{t+1})$, where $\mathbf{D}$ denotes the decoder 
(see Eq.~(\ref{eq:encoder-decoder})). 
It is important to note here that $\mathbf{h}_{t}$ also contains information about $\mathbf{h}_{t-1}$.
Thus, the predicted latent vector $\hat{\Bell}_{t+1}$ at time step $t$ not only depends on $\Bell_{t}$, but also on $\Bell_{t-1}$ as per Eq.~\eqref{eq:gru-a}, and in the case of image data only, no sensor conditioning is present.

When the vehicle's state is concatenated with the latent vector space, the concatenated vector is multiplied with the fully connected layer with weights $W_{fc}$. There are two reasons to use fully connected layers before and after the \gls{gru} cell: 1) The fully connected layer acts as a normalization layer when the latent vector $\mathbf{e}_t$ is concatenated with $\mathbf{s}_t$ and scale concatenated vector appropriately. 2) The fully connected layer after the \gls{gru} again serves the purpose of scaling the $h_t$ appropriately for the decoder.
To summarize, we train  the \gls{KARNet} network to model the following transitions:

\begin{enumerate}
	\item[a)] 
	      In case of image-based latent representation only (Fig.~\ref{fig:combined_architecture_all}), omitting the vehicle's state $\mathbf{s}_t$, 
       the model is trained to represent the transition $T(\hat{\Bell}_{t+1}|\Bell_{t}, \mathbf{h}_{t})$.
	          
	\item[b)] 
	      In case of an augmented latent representation (image + vehicle's state, Fig.~\ref{fig:combined_architecture_all}), the model is trained to represent the transition $T(\hat{\Bell}_{t+1}|\Bell_{t}, \mathbf{s}_{t}, \mathbf{h}_{t})$.
\end{enumerate}

It is important to note that the modeled recurrent network transition is
different from the one in World Models (WM)~\cite{ha2018recurrent}. 
In WM, the \gls{mdnrnn} action-conditioned state transition is defined as $T(\hat{\Bell}_{t+1}|{\Bell}_{t}, \mathbf{h}_{t}, \mathbf{a}_{t})$, where $\Bell_{t}$ and $\hat{\Bell}_{t+1}$ are the present and predicted latent states, 
$\mathbf{a}_{t}$ is the action. 
On the other hand, we do not condition the latent vector on actions, i.e., \gls{KARNet} learns $T(\hat{\Bell}_{t+1}| \Bell_{t}, \mathbf{h}_{t})$ or $T(\hat{\Bell}_{t+1}|\Bell_{t}, \mathbf{s}_{t}, \mathbf{h}_{t})$, in case the vehicle state is used alongside with camera images. 
The main reasoning is that the actions can be inferred from the model's state $\mathbf{s}_{t}$.
However, action information can be added, if necessary. 
In that case, the \gls{KARNet} model represents the transition $T(\hat{\Bell}_{t+1}|\Bell_{t}, \mathbf{s}_{t}, \mathbf{h}_{t}, \mathbf{a}_{t})$.

\subsection{Error-State Extended Kalman Filtering (EEKF)}

\begin{figure}[th!]
	\centering
	\includegraphics[width=0.75\columnwidth]{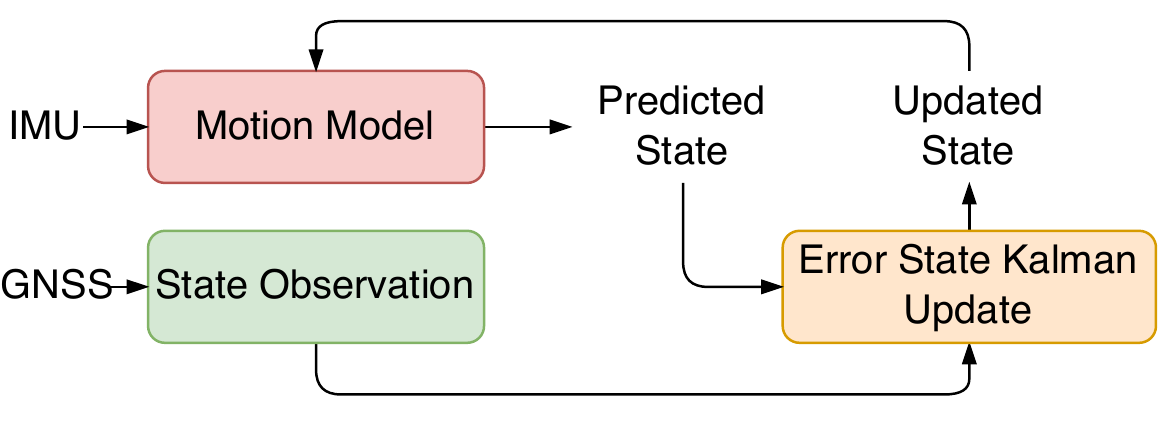}
	\caption{Error-state extended Kalman filtering approach for estimating the vehicle's state using IMU and Global navigation satellite system (GNSS).}
	\label{fig:Kalman}
\end{figure}

We use an error-state extended Kalman filter to estimate the state of the vehicle and integrate the states with \gls{KARNet}. 
Note that any filtering algorithm can be used to estimate the vehicle state, and the state vector $\mathbf{s}_t$ can be expanded to include orientation. 
The overview of the error-state extended Kalman filtering procedure is shown in Fig.~\ref{fig:Kalman}. 
The motion model uses IMU sensors to predict the next state using the vehicle dynamics. 
The global navigation satellite system (GNSS) is used to update the state using Kalman filtering.
Also, the position and velocity quantities are normalized using the maximum velocity and maximum size of the map.

Without loss of generality, we assume that the motion model is given by equation~\eqref{eq:motion_model}
\begin{subequations}
	\begin{align}      \label{eq:motion_model}
		  & \mathbf{p}_{t} = \mathbf{p}_{t-1}+\mathbf{v}_{t-1} \Delta t+\frac{1}{2}\left(\mathbf{R}_{t-1}\left(\mathbf{a}_{m, t-1}\right)+\mathbf{g}\right) \Delta t^2, \\
		  & \mathbf{v}_{t} = \mathbf{v}_{t-1}+\left(\mathbf{R}_{t-1}\left(\mathbf{a}_{m, t-1}\right)+\mathbf{g}\right) \Delta t,                                        \\
		  & \mathbf{q}_t = \mathbf{q}_{t-1} \otimes \mathbf{q}_{t-1}\left\{\left(\boldsymbol{\omega}_{m, t}\right) \Delta t\right\},                                    
	\end{align}
\end{subequations}
where $\mathbf{p}_t, \mathbf{v}_t$, and $\mathbf{q}_t$ are position, velocity, and orientation (the latter represented by the quaternion $\mathbf{q}$). 
Here
$\mathbf{R}_t$ is the rotation matrix associated with the quaternion $\mathbf{q}_t$, $\Delta t$ is the sampling time, $\mathbf{g}$ is the gravitation vector, and $\mathbf{a}_{m, t}$ and $\mathbf{\omega}_{m, t}$ are acceleration and orientation measured from the IMU sensor. 
We have neglected the bias in the IMU sensor readings to simplify the analysis.
Note that even though we have used the orientation ($\mathbf{q}$) in the motion model, for the \gls{KARNet} Kalman state $\mathbf{s}_t$, we have used only position and velocity components not the orientation component, i.e., $\mathbf{s}_t=\mqty[p_x & p_y & v_x & v_y]$.
Since the filtering procedure includes correction and prediction states, we will use $\mathbf{s}^c_t$ to represent the corrected state from the previous time step.
For the prediction step, we  use $\mathbf{s}^p_t$.
The state uncertainties are propagated according to equation~\eqref{eq:uncertainty}
\begin{equation}
	\label{eq:uncertainty}
	{\mathbf{P}}_t=\mathbf{F}_{t-1} \mathbf{P}_{t-1} \mathbf{F}_{t-1}^{\top}+\mathbf{L}_{t-1} \mathbf{Q}_{t-1} \mathbf{L}_{t-1}^{\top},
\end{equation}
where $\mathbf{F}_t$ and $\mathbf{L}_t$ are the motion model Jacobians' with respect to state and noise, respectively, and $\mathbf{Q}_t$ is the noise covariance matrix at time step $t$.
Once the measurements are available, the filter correction step is carried out according to 
\begin{subequations} 
	\begin{align}
		  & \mathbf{K}_t=\mathbf{P}_t \mathbf{H}_t^{\top}\left(\mathbf{H}_t \mathbf{P}_t \mathbf{H}_t^{\top}+\mathbf{V}\right)^{-1}, \\
		  & \delta \mathbf{s}_t = \mathbf{K}_t\left(\mathbf{y}_t-\mathbf{s}_t\right),                                                \\
		  & \mathbf{P}_t = (\mathbf{I}-\mathbf{K}_t \mathbf{H}_t) \mathbf{P}_t,                                                      
	\end{align}
\end{subequations}
where $\mathbf{y}_t$ is the measurement from GNSS, $\mathbf{H}_t$ is the measurement model Jacobian, and $\mathbf{V}$ is the measurement covariance. 
The states are corrected according to $\mathbf{s}^c_t = \mathbf{s}_t + \delta \mathbf{s}_t$. 
After this step, we can use the model (from equation~\eqref{eq:motion_model}) to propagate the states ($\mathbf{s}^p_t$) forward. 
Note that the propagated states are available at time step $t$ using the motion model.
Thus, at each time instance, we have two sets of states, i.e., corrected state using filtering ($\mathbf{s}^c_t$) and predicted states ($\mathbf{s}^p_t$) using the motion model.
The following section will discuss two ways of integrating the corrected and predicted states with \gls{KARNet}.

\subsection{Early and Late-Fusion}

\begin{figure}[th!]
	\centering
	\includegraphics[width=0.75\columnwidth]{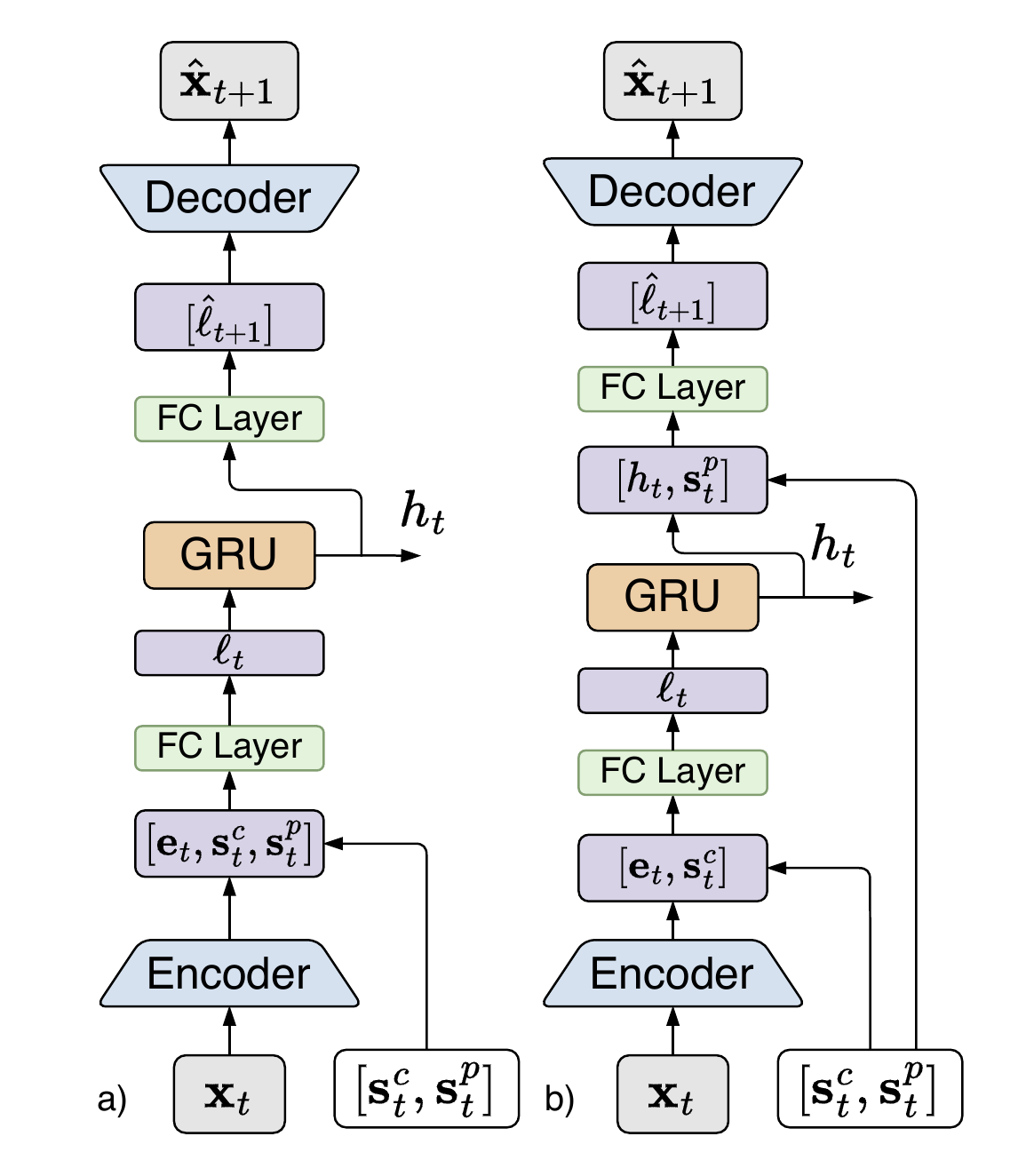}
	\caption{Early and late-fusion approach to integrating the vehicle's state vector with latent vector from image space.}
	\label{fig:fusion}
\end{figure}

Given the corrected ($\mathbf{s}^c_t$) and predicted states ($\mathbf{s}^p_t$), we explored two ways of integrating the states with the \gls{KARNet} as shown in Fig.~\ref{fig:fusion}, namely, early-fusion and late-fusion. 

In early fusion, both the corrected state $\mathbf{s}^c_t$ and predicted state $\mathbf{s}^p_t$ are directly concatenated with the latent vector $\mathbf{e}_t$ before the \gls{gru} block. 
The concatenated vectors are scaled to the appropriate size using fully connected layers. 
The rationale for using both the states ($\mathbf{s}^c_t$ and $\mathbf{s}^p_t$) is to inform the network about the present and predicted state of the vehicle early on and facilitating the prediction of the next image frame. 

In the late-fusion, the corrected state $\mathbf{s}^c_t$ is concatenated with $\mathbf{e}_t$, while the predicted state $\mathbf{s}^p_t$ is concatenated with the output of the \gls{gru} $h_t$. 
The reasoning is that $\mathbf{s}^p_t$ captures the future state and thus might help capture the next frame information. 
Again, the fully connected layer scales the concatenated vector appropriately for the decoder network.

\subsection{Training}   \label{subs:training}

For the loss function, we have used \gls{msssim}~\cite{wang2003multiscale} for image prediction to preserve the reconstructed image structure and also serve as some form of regularization.
In our experiments, 
using \gls{mse} alone resulted in mode collapse and blurry reconstructions. 
SSIM is a similarity measure that compares structure, contrast, and luminance between images. 
\gls{msssim} is a generalization of this measure over several scales~\cite{wang2003multiscale}, as follows
\begin{equation}
	L_{\text{\tiny{MS-SSIM}}} = \left[l_M(\mathbf{x}, \mathbf{y})\right]^{\alpha_M} \prod_{j=1}^{M} \left[c_j(\mathbf{x}, \mathbf{y})\right]^{\beta_j} \left[s_j(\mathbf{x}, \mathbf{y})\right]^{\gamma_j},
	\label{eq:ms-ssim}
\end{equation}
where $\mathbf{x}$ and $\mathbf{y}$ are the images being compared,  $c_j(\mathbf{x}, \mathbf{y})$ and $s_j(\mathbf{x}, \mathbf{y})$ are the contrast and structure comparisons at scale $j$, and the luminance comparison $l_M(\mathbf{x}, \mathbf{y})$ (shown in Eq.~\eqref{eq:ms-ssim-comparisons}) is computed at a single scale $M$, and
$\alpha_M$ and $\beta_j$, $\gamma_j$ ($j=1,\ldots,N$) are weight parameters that are used to adjust the relative importance of the aforementioned components, i.e., contrast, luminance, and structure. These parameters are left to the default implementation values as follows
\begin{subequations}    \label{eq:ms-ssim-comparisons}
	\begin{align}
		l(\mathbf{x}, \mathbf{y})  & = \frac{2\mu_x \mu_y + C_1}{\mu_x^2 + \mu_y^2 + C_1},             \\
		c(\mathbf{x},  \mathbf{y}) & = \frac{2\sigma_x \sigma_y + C_2}{\sigma^2_x + \sigma^2_y + C_2}, \\
		s(\mathbf{x}, \mathbf{y})  & = \frac{\sigma_{xy} + C_3}{\sigma_x\sigma_y + C_3}.               
	\end{align}
\end{subequations}
In (\ref{eq:ms-ssim-comparisons}), $\mu$ and $\sigma$ are the mean and the variance of the image and are treated as an estimate 
of the luminance and contrast of the image. 
The constants $C_1,C_2,C_3$ are given by $C_1=(K_1L)^2$, $C_2=(K_2L)^2$, $C_3=C_2/2$, where $L$ is the dynamic range of the pixels and $K_{1}, K_2$, are two scalar constants. 
Additionally, not only the final predicted variable is used in the loss function, but all intermediate predictions of the recurrent network are also used to improve continuity in the latent space.

\subsection{Action Prediction Network}

\begin{figure}[t!] 
	\centering
	\includegraphics[width=\linewidth]{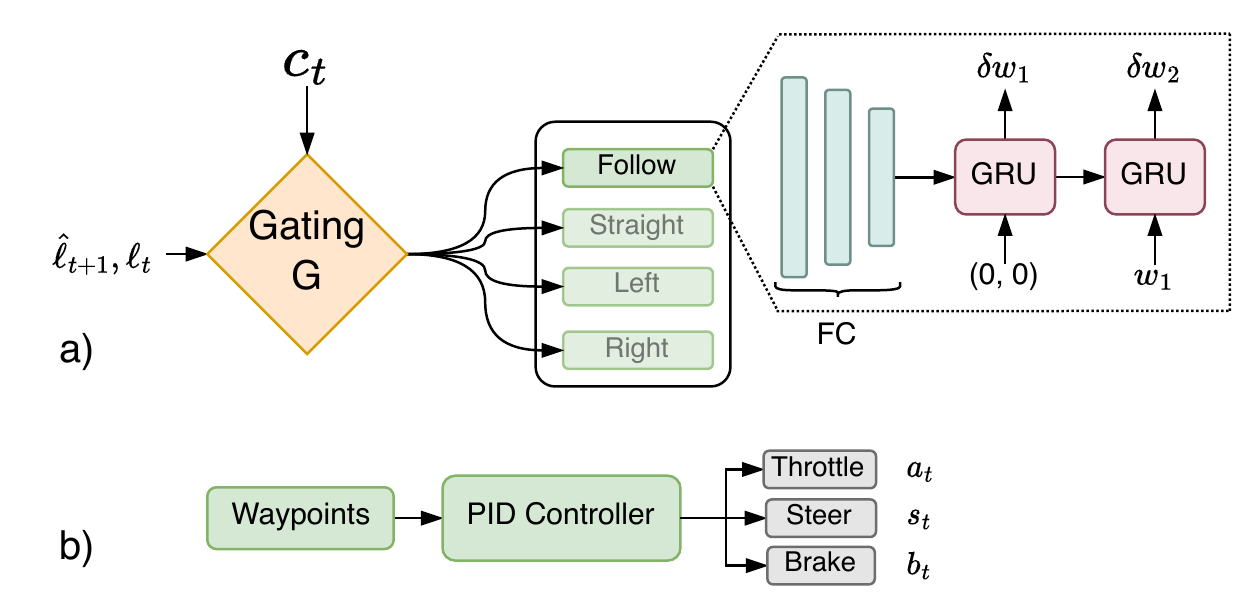}
	\caption{a) Action Prediction network architecture is used for predicting the waypoints. Each sub-network (Follow, Straight, Left, Right) consists of 3 layers of fully connected (FC) and a two-time step \gls{rnn} network, which is used to predict the two waypoints by taking the vehicle's present location as the origin. b) A PID controller converts the waypoints to throttle/acceleration, steer, and brake.}
	\label{fig:action_network}
\end{figure}

For the action prediction network, we use the \gls{KARNet} as a feature extractor. 
In our case, the features are the latent representations.
Note that the \gls{KARNet} weights are frozen and only the action prediction network weights are trained.
Training the prediction model (i.e., \gls{KARNet}) separately from the action model has parallels in neuroscience where the pre-motor areas can be activated without the resulting motor function \cite{hesslow2012current}.
The control command $c_{t}$ is introduced to handle goal-oriented behavior starting from an initial location to reach the final destination (similar to \cite{liang2018cirl}). 
The command $c_{t}$ is a categorical variable that controls the selective branch activation via the gating function $G(c_{t})$, where $c_{t}$ can be one of four different commands, i.e., follow the lane (Follow), drive straight at the next intersection (Straight), turn left at the next intersection (TurnLeft), and turn right at the next intersection (TurnRight). 
Four policy branches are specifically learned to encode the distinct hidden knowledge for each case and thus selectively used for action prediction. 
Each policy branch is an auto-regressive waypoint network implemented using GRUs \cite{chitta2022transfuser} (Fig~\ref{fig:action_network}).
The first hidden vector for the GRU network is calculated using the latent vectors from \gls{KARNet} as $h_0=\mathrm{FC}(\hat{\Bell}_{t+1}, \Bell_{t})$ where $\mathrm{FC}$ is a fully connected neural network of output size $64$.
The input for the GRU is the position of the previous waypoint. 
The output of the GRU network is the difference between the current and next waypoint, i.e., $w_i = w_{i-1} + \delta w_i$ where $w_{i-1}$ is the input to the GRU and $\delta w_i$ is the prediction. 
The waypoints are predicted by taking the car's current position as reference, hence $w_0 = (0, 0)$.

\subsection{Separate and Ensemble Training} 

When pre-training the autoencoder separately on individual images, no temporal relations are captured in the latent space. 
This can potentially impact the quality of predictions produced by the \gls{rnn} network. 
Since both the autoencoder and the \gls{rnn} parts are trained together, we did not train the combined architecture from scratch but only pre-trained the autoencoder and then fine-tuned it as part of a bigger network. 
This resulted in significantly less time spent when training the combined architecture.

\subsection{Simulated Data}

In order to learn good latent state representations from image data, a significant amount of data is required. 
Therefore, the frontal, side, and rear camera images are also included for regularizing and increasing data diversity (e.g., the side camera image becomes similar to the frontal camera image once the vehicle makes a right-angle turn). 
It is important to note that only the autoencoder part was pre-trained on all camera feeds, and the ensemble network only used the frontal camera. 
This proved helpful when using real-world datasets with misaligned data, i.e., with the absence of a global shutter and mismatching data timestamps.

\begin{figure}[th!]
	\centering
	\includegraphics[width=0.6\columnwidth]{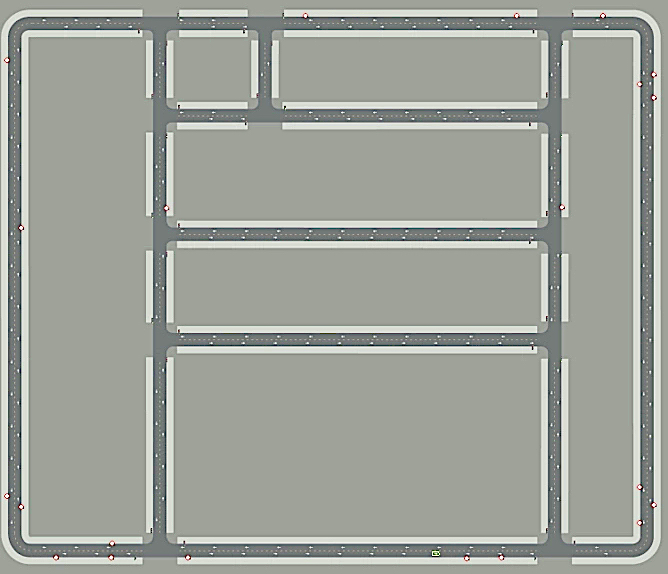}
	\caption{Map of the town used in CALRA for data generation.}
	\label{fig:town}
\end{figure}

The training data was generated using the CARLA \cite{dosovitskiy2017carla} simulator. 
A total of 1.4M time steps were generated using random roll-outs (random starting and goal points on the map shown in Fig.~\ref{fig:town}) utilizing the internal CARLA vehicle autopilot. 
The simulated data includes four-directional camera images (front/left/right/rear) along with their corresponding semantic segmentation, IMU, waypoints for navigation and other sensor data (speed, steering, LIDAR, GNSS, etc.), desired control values, and additional experimental data for auxiliary tasks such as traffic light information.

\section{Experiments}

We considered two cases to demonstrate the efficacy of the proposed architecture.
First, we considered an \gls{il} and \gls{rl} case in which the task is to predict the waypoints given the front view camera for the simulated scenario in the CARLA simulator.
In the second case, we predict the steer, acceleration, and brake actions from a real dataset.

\subsection{Metrics}  \label{sec:metrics}

We use two metrics to assess the driving behavior of the agent: Route Completion and Infractions per km. 
The bench-marking is run in the Town01 map (Fig~\ref{fig:town}) of CARLA with a combination of weather and daylight conditions not seen during the training. 
The agent is allowed the drive between two randomly chosen points (an average distance of 1.5 km) on the map with 20 vehicles and 50 pedestrians.
The metrics are averaged over five runs.
We terminate the episode and calculate the metric if the agent goes off-road or collides with other agents/static elements.

\begin{itemize}
	\item Route Completion (RC): is the percentage of the route completed given the start and finish location. 
	      Note that depending on the start and finish locations, the total route distance might differ for each experiment run. 
	      Hence, while reporting the metric, we take the average.
	\item Infractions per km (IN/km): the infraction (violations) considered in this study are collision with vehicles, pedestrians, static elements, lane deviations, and off-road driving. 
	      We count the infractions for the given route and normalize by the total number of km driven as 
	      \begin{equation}
	      	\begin{aligned}
	      		\text{Infraction per km} = \frac{\sum \# \text{Infractions}}{K} 
	      	\end{aligned}
	      \end{equation}
	      where $K$ is the total distance traveled in the given route. 
\end{itemize}

\subsection{Reinforcement Learning}  \label{sec:RLexp}

We carried out the data collection and reinforcement learning training in the CARLA environment~\cite{dosovitskiy2017carla}. 
We modified the OpenAI Gym wrapper \cite{chen2020interpretable} for the CARLA simulator and used the customized version as an interface for the environment required for generating data for our learning algorithms.
The reward function used for training is given as
\begin{equation}
	\begin{aligned}
		r = 200 \, r_{\rm collision} & + v_{\rm lon} + 10\frac{S}{S_\text{des}} \, r_{\rm fast}   \\
		                             & + 40 \, r_{\rm out} -5 \alpha^2 + 0.2 \, r_{\rm lat} -0.1, 
	\end{aligned}
\end{equation}
where $r_{\rm collision}$ is set to $-1$ if the vehicle collides; else, it is set to $0$; $v_{\rm lon}$ is the magnitude of the
projection of the vehicle velocity vector along the direction connecting the vehicle and the nearest next waypoint;
$r_{\rm fast}$ is $-1$ when the car is faster than the desired speed; else, it is 0; $r_{\rm out}$ is $-1$ when the vehicle is out of the lane; $\alpha$ is the steering angle; $r_{\rm lat}$ refers to the lateral acceleration = $\alpha v^2$; the constant $-0.1$ is to make sure that the vehicle does not remain standstill. 
The penalty $r_{\rm fast}$ is either 0 or $-1$, which acts as a constraint for over-speeding.
Note that the coefficient for $r_{\rm out}$ is relatively large, as going out-of-lane causes a termination of the episode. 
Thus, the negative reward is only experienced once during an episode. 
Furthermore, a weight factor derived from normalizing the vehicle's speed $S$ by the desired speed $S_{\rm des} (30~\rm Km/h)$ is added to $r_{\rm fast}$; otherwise, the reward function would encourage a full-throttle policy. 
The action space for the agent is predicting the waypoints (continuous $x$ and $y$ position) instead of steering, acceleration, and braking.
Note that the waypoints are converted to steer, acceleration, and brake using a PID controller, which does not contain any trainable parameters.
Since the action space is continuous and deterministic, we have used the DDPG~\cite{mnih2013playing} algorithm for training the RL model. 
The hyper-parameters for the DDPG model are shown in Table~\ref{tab:dqn_hyper_parameters}.

\begin{table}
	\begin{minipage}{.45\linewidth}
		\centering
		\setlength\tabcolsep{1.25pt}
		\renewcommand{\arraystretch}{1.10}
		\caption{Learning Hyperparameters}
		\vspace{-8pt}
		\begin{tabular}[t]{lc}
			\hline
			\textbf{Hyperparameter}   & \textbf{Value} \\
			\hline
			Input image size          & 256x256        \\
			Autoencoder latent size   & 128            \\
			\gls{gru} hidden size     & 128            \\
			RNN time-steps            & 4              \\
			Optimizer                 & Adam           \\
			Learning rate (scheduled) & $10^{-3}$      \\
			Batch size                & 64             \\
			\hline
		\end{tabular}
		\label{tab:learning_hyper_parameters}%
	\end{minipage}%
	\hfill%
	\begin{minipage}{.45\linewidth}
		\centering
		\setlength\tabcolsep{1.25pt}
		\renewcommand{\arraystretch}{1.10}
		\caption{DDPG Hyperparameters}
		\vspace{-8pt}
		\begin{tabular}[t]{lc}
			\hline
			\textbf{Hyperparameter} & \textbf{Value}    \\
			\hline
			Training steps          & 500000            \\
			Buffer size             & 5000              \\
			Optimizer               & Adam              \\
			Actor Learning rate     & $5\times 10^{-3}$ \\
			Critic Learning rate    & $5\times 10^{-3}$ \\
			Batch size              & 32                \\
			N-step Q                & 1                 \\
			\hline
		\end{tabular}%
		\label{tab:dqn_hyper_parameters}%
	\end{minipage} 
\end{table}

\subsection{Evaluation using Real World Images}

We evaluated  the \gls{KARNet} performance on the Udacity dataset~\cite{udacity2018} that consists of real-world video frames taken from urban roads. 
The Udacity dataset posed two main problems, which made the direct use of the \gls{KARNet} difficult. 
First, it does not have a global shutter, resulting in a camera and sensor timestamp mismatch.
To remedy this, front camera timestamps were chosen as reference, and the closest images and sensor data were selected in relation to that timestamp.
However, running the extended Kalman filtering procedure might not produce correct state estimation corresponding to the image captured from the front camera. 
Hence, we directly used the IMU and GNSS measurements in an early-fusion manner without estimating the states of the car using extended Kalman filtering. 
The rationale is that IMU/GNSS information indirectly captures the car's state and thus should improve the performance of the \gls{KARNet}. 

\begin{figure}[th!]
	\centering
	\includegraphics[width=\columnwidth]{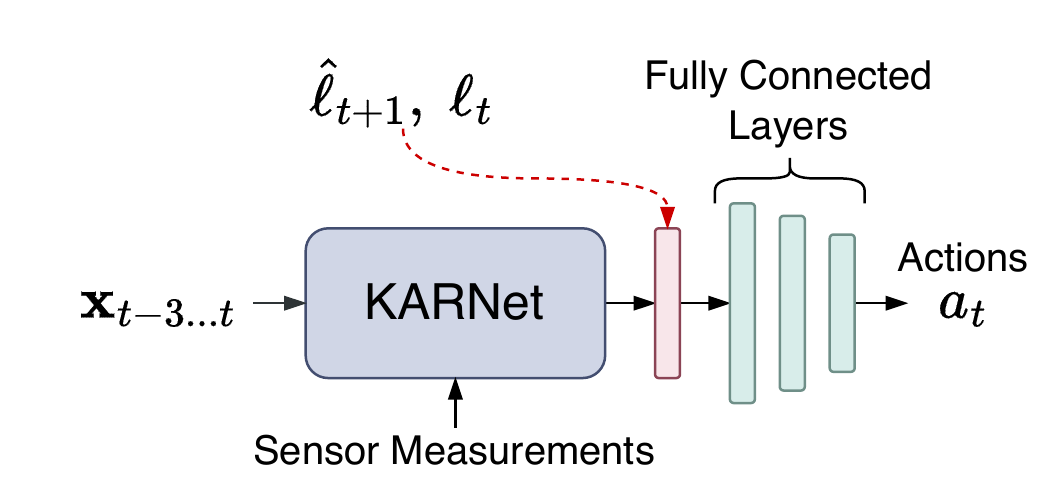}
	\caption{For imitation learning, \gls{KARNet} and \gls{wm} are not conditioned on history of actions. However, for reinforcement learning, both the models are conditioned on the history of actions.}
	\label{fig:experimental-setup}
\end{figure}

Second, we cannot extract the performance metrics (see Section \ref{sec:metrics}) from an offline dataset (i.e., the learned agent cannot interact similarly to a simulator). 
Instead, we can cast the agent's performance as a classification problem that aims to classify the correct action given the image. 
For this purpose, we replaced the waypoint prediction network with a classifier network, as shown in Fig.~\ref{fig:experimental-setup}. 

In order to match the simulated dataset scenario and structure, the extracted images were resized to $256\times 256$ dimensions and converted to grayscale. 
We post-processed throttle and steering values that were extracted from the rosbag file.
The post-processing step is as follows: The steering angle is discretized into $[-0.2, 0.0, 0.2]\ \text{rad}$ and the acceleration command is discretized into $[-3, 0.0, 3]\ \text{m/s}^2$ resulting in a nine-class classification problem from different combinations of the discrete steering and acceleration values.
Discretization of the action space is justifiable from a practical perspective, 
since in autonomous driving applications, the high-level decision policy derived by imitation learning, reinforcement learning, game theory, or other decision-making techniques is combined with low-level trajectory planning and motion control algorithms \cite{nageshrao2019autonomous}.
For training, we used mean cross-entropy loss between the predicted and the autopilot actions. 

\begin{table}[ht!]
	\renewcommand{\arraystretch}{1.25}
	\centering
	\caption{Controller structure for evaluation using real-world images}
	\label{tab:controller_architecture}
	\begin{tabular}{ccc}
		\hline
		\textbf{Layer} & \textbf{Input Size} & \textbf{Output Size} \\ 
		\hline
		FC1 + ReLU     & 256                 & 128                  \\ 
		FC2 + ReLU     & 128                 & 128                  \\ 
		FC3 + ReLU     & 128                 & 64                   \\ 
		FC4 + ReLU     & 64                  & 9                    \\ \hline
	\end{tabular}
\end{table}

The controller architecture for the imitation learning task
consists of a simple four-layer \gls{mlp} that gradually reduces the dimensionality of the latent space onto the action space (see Fig.~\ref{fig:experimental-setup}). 
The structure of the controller is given in Table~\ref{tab:controller_architecture}. 
The controller network uses the stacked previous and predicted latent vectors, so the input tensor size 
is twice the chosen latent size $\mathbf{a}_{} = \mathbf{C}([\hat{\Bell}_{t+1}, \, \Bell_{t}])$, e.g., for a latent size of $128$, the size of the input tensor to the imitation learning network is $256$.
The training was performed on a computer equipped with a GeForce RTX3090 GPU, Ryzen 5950x CPU, and 16GB of RAM. 
Also, the 300K timesteps were split into 70\% training, 15\% validation, and 15\% testing data.
Note that this split is for the Udacity dataset.
The code implementation\footnote{\texttt{\url{https://github.com/DCSLgatech/KARNet}}} uses Pytorch with the Adam optimizer.

\section{Results}

We first present the ablation study of the \gls{KARNet} followed by imitation learning, reinforcement, and real-world image analysis.

\subsection{Ablation Analysis}

We performed ablation studies to select the best network (i.e., \gls{KARNet} without any state vector augmentation) configuration and hyper-parameters. Ablation studies were performed on the simulated dataset from the CARLA simulator. The parameters of the ablation study are summarized in Table \ref{tab:ablation_parameters}, and selected parameters are highlighted in bold. 

\begin{table}[!ht]
	\renewcommand{\arraystretch}{1.5}
	\centering
	\caption{Ablation Studies Overview}
	\begin{tabular}{cccc}
		\hline
		\textbf{Parameters} & \textbf{Latent Size}    & \textbf{RNN Unit}    & \textbf{Loss Function}  \\
		\hline
		\textbf{Range}      & [64, \textbf{128}, 256] & [LSTM, \textbf{GRU}] & [MSE, \textbf{MS-SSIM}] \\
		\hline
	\end{tabular}
	\label{tab:ablation_parameters}
\end{table}%

First, we tested various latent sizes for the autoencoder and hidden vector sizes for the \gls{rnn}. 
While reducing the latent size did address the over-fitting problem, reducing the size further led to low representation capacity (summarized in Fig.~\ref{fig:ablation_ae}). 
Hence, we chose a latent size of 128, providing a good trade-off between the representation capacity and the number of trainable parameters.
Second, during the initial experiments, \gls{lstm} was replaced in favor of \gls{gru}, as the former showed heavy over-fitting behavior (see Fig.~\ref{fig:ablation_lstm_gru}). 
Lastly, we selected \gls{msssim} for the image reconstruction loss as \gls{mse} often resulted in mode collapse and blurry image reconstruction since it did not impose any structural constraints on the decoded image. 
 
\subsection{Separate and Ensemble Training}

Recall that the \gls{KARNet} model consists of an autoencoder with a \glsreset{rnn} \gls{rnn} trained together. 
To speed up the training process, we pre-trained the autoencoder and fine-tuned the weights as part of the whole network, i.e., \gls{ae} + \gls{rnn} (see Fig.~\ref{fig:combined_architecture_all}). 
First, we present the results of the pre-training procedure.

\begin{figure}[ht]
	\centering
	\includegraphics[width=0.9\columnwidth]{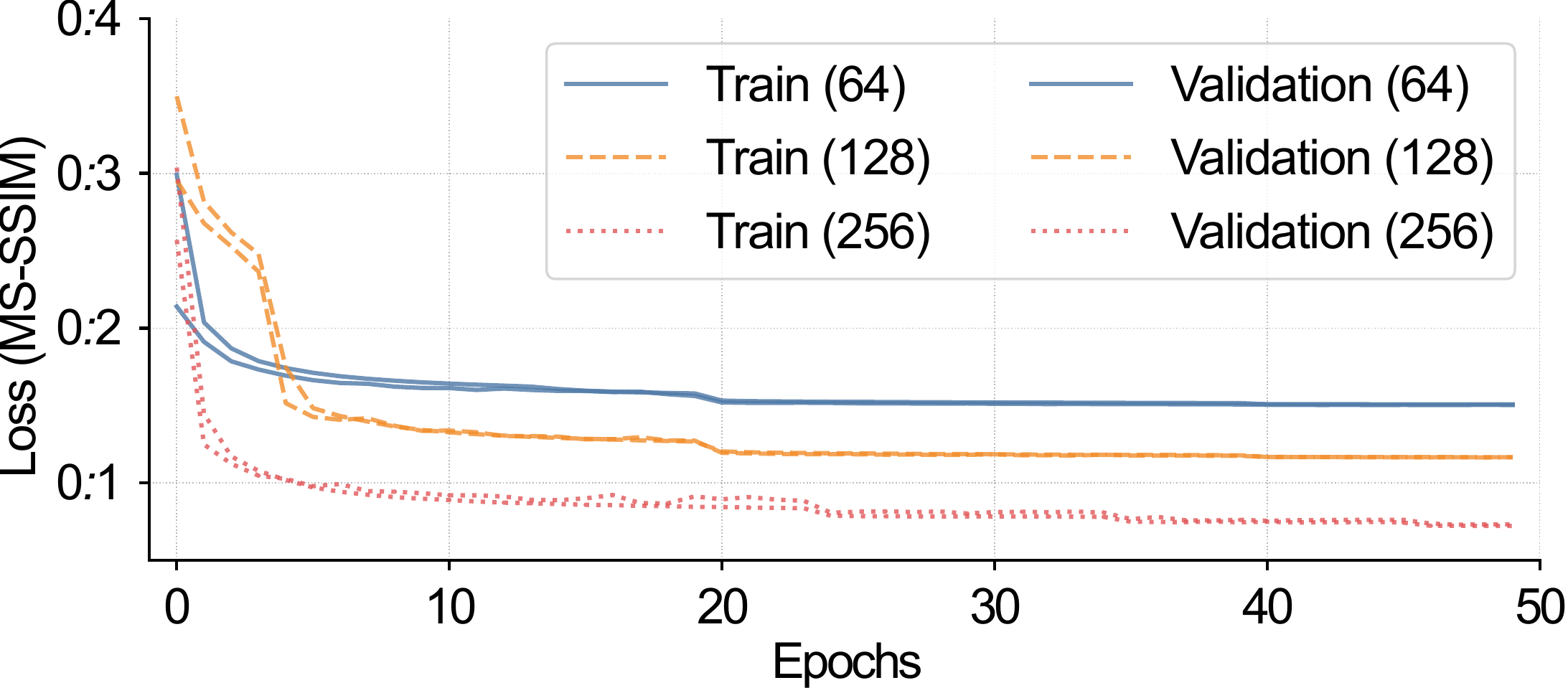}
	\caption{MS-SSIM loss for autoencoder reconstructions of various latent variable sizes. Note that the range for the \gls{msssim} loss is $[0, 1]$, where 0 corresponds to perfect structure match, while 1 shows complete mismatch, accordingly. A nearly linear relation between the latent variable size and final reconstruction quality can be observed. }
	\label{fig:ablation_ae}
\end{figure}

\begin{figure}[ht]
	\centering
	\includegraphics[width=0.9\columnwidth]{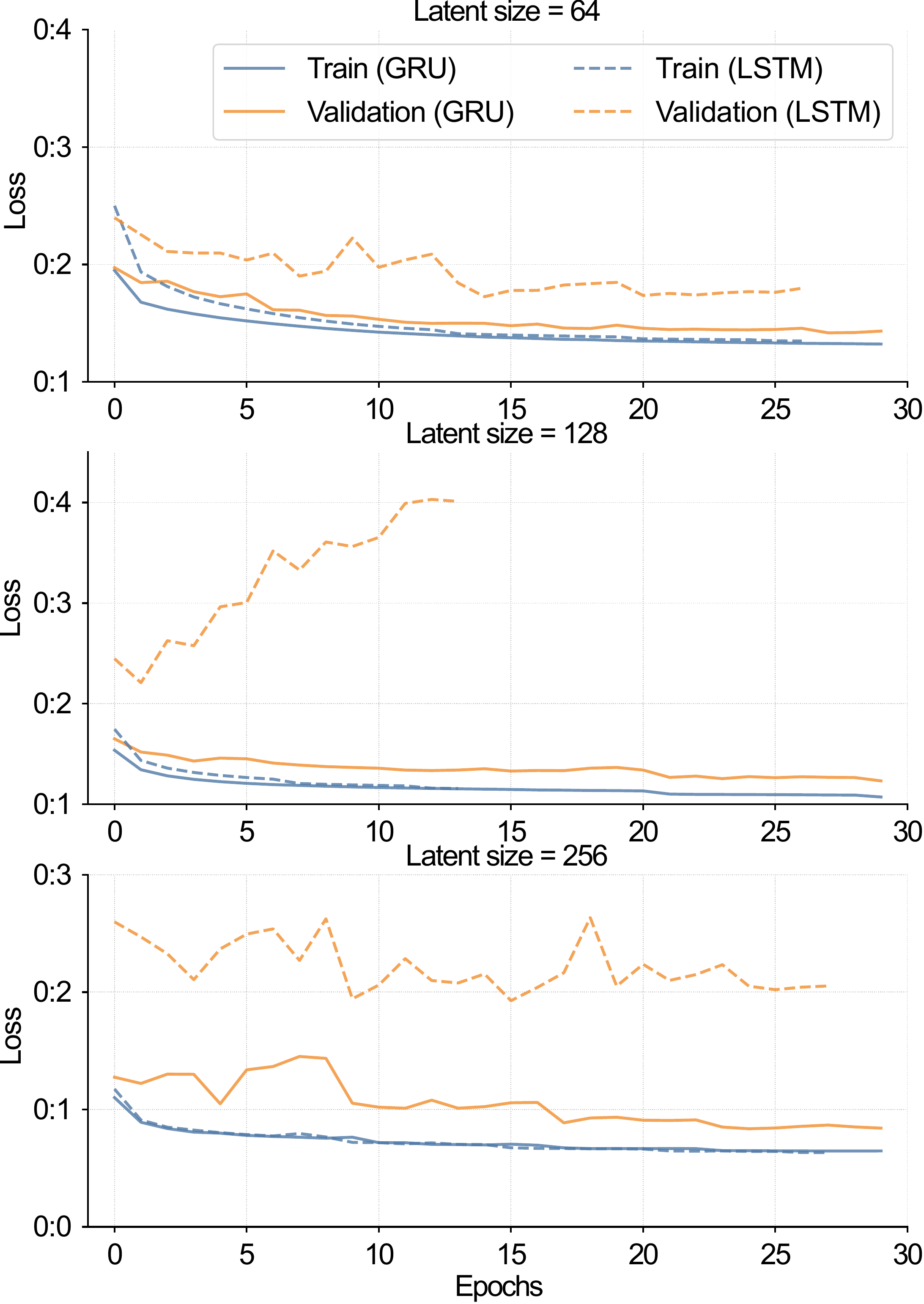}
	\caption{\gls{lstm} vs. GRU varying latent variable size training and validation curves. Note that \gls{lstm} training often triggered early stopping criteria (no decrease in validation loss, even with learning rate scheduling and showed a high tendency to overfit compared to \gls{gru}. Both \gls{rnn} architectures were trained under the same set of hyperparameters.}
	\label{fig:ablation_lstm_gru}
\end{figure}

\begin{figure}[ht!]
	\centering
	\subfigure[Source (CARLA)]{\includegraphics[width=0.45\columnwidth]{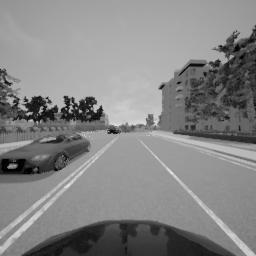}}
	\subfigure[Reconstructed (CARLA)]{\includegraphics[width=0.45\columnwidth]{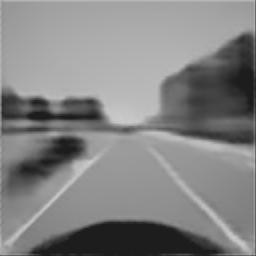}}
		
	\subfigure[Source (Udacity)]{\includegraphics[width=0.45\columnwidth]{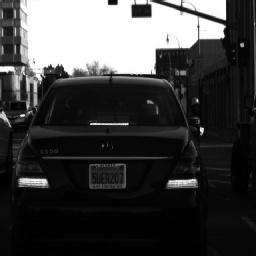}}
	\subfigure[Reconstructed (Udacity)]{\includegraphics[width=0.45\columnwidth]{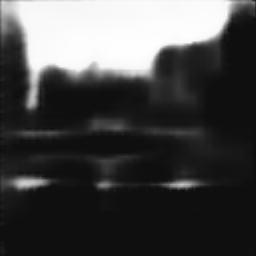}}
	\caption{Autoencoder reconstruction examples for simulated data and real-world data.}
	\label{fig:results_ae_reconstruction}
\end{figure}

Figure~\ref{fig:ablation_ae} (latent size 128) shows the training and validation curve for pretrained models, suggesting no over-fitting at the pre-training stage. 
Example reconstructions for simulated and real data after the pre-training stage are shown in Fig.~\ref{fig:results_ae_reconstruction}.  

\begin{figure*}[!ht]
	\centering
	\subfigure[Source ]{\includegraphics[width=0.19\textwidth]{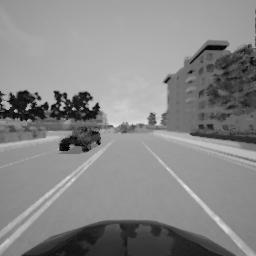}}
	\subfigure[Prediction (LSTM, 128)]{\includegraphics[width=0.19\textwidth]{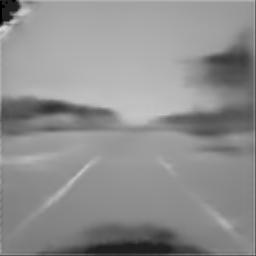}}
	\subfigure[Prediction (GRU, 128)]{\includegraphics[width=0.19\textwidth]{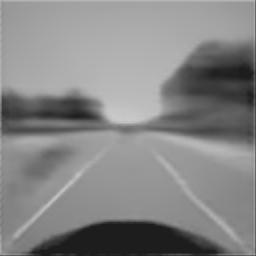}}
	\subfigure[Prediction (LSTM, 64)]{\includegraphics[width=0.19\textwidth]{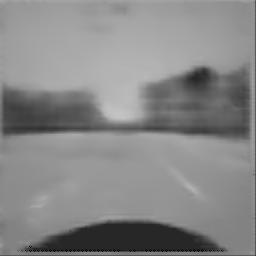}}
	\subfigure[Predcition (GRU, 64)]{\includegraphics[width=0.19\textwidth]{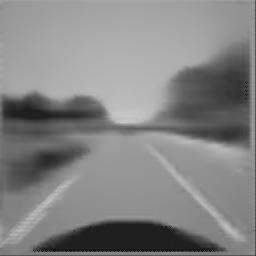}}
	\caption{Comparison of LSTM and GRU reconstruction with different latent vector sizes. Note that the reconstructed images are from the latent vector predicted (next time-step) using LSTM/GRU RNN cells.}
	\label{fig:results_ablation_ae_comparison_2}
\end{figure*}

As seen in Fig.~\ref{fig:results_ae_reconstruction}, the autoencoder can adequately capture all necessary parts of the source image. 
Nonetheless, it is essential to note that our main aim was to learn latent dynamics rather than ideal image reconstruction.

After pre-training the autoencoder, we combined it with the \gls{rnn} and trained them 
together without freezing the weights of the autoencoder.
The training losses for the \gls{KARNet} with \gls{gru} are shown in Fig.~\ref{fig:ablation_lstm_gru}. 
The \gls{gru} performed better than the \gls{lstm} in terms of final loss value and overfitting. 
The slight overfit in the image reconstruction appears due to the dataset size. 
Qualitatively, Fig.~\ref{fig:results_ablation_ae_comparison_2} shows image reconstructions from the predicted latent space.
The quantitative differences in the learned latent variables become more apparent when the action network is added, as shown in the following subsection.

\subsection{Imitation Learning (CARLA)}

Before running the \gls{rl} task, we pre-trained the action network model using imitation learning and offline data.
Table~\ref{tab:results_imitation_latent} shows the imitation learning results where the action network is trained in a supervised manner. 
Both late-fusion and early fusion \gls{KARNet} perform better than the \gls{KARNet} without the state (from EEKF estimation) augmentation.
\begin{table}[htbp]
	\renewcommand{\arraystretch}{1.5}
	\centering
	\caption{Bench-marking results.}
	\begin{tabular}{p{30mm}cc}
		\hline
		\textbf{Model}               & \textbf{Route Completion} & \textbf{No. of Infractions} \\
		\hline
		\textbf{Without EEFK States} & 42.26                     & 40.36                       \\
		\textbf{Late Fusion}         & 45.29                     & 14.5                        \\
		\textbf{Early Fusion}        & \textbf{55.75}            & \textbf{11.2}               \\
		\hline
	\end{tabular}%
	\label{tab:results_imitation_latent}%
\end{table}%
Moreover, early fusion \gls{KARNet} performs significantly better than late fusion. 
A probable reason for the difference between the late and early fusion is that the GRU block in the \gls{KARNet} (see Fig.~\ref{fig:fusion}) does not have access to the predicted states ($\mathbf{s}_{t}^p$) to learn the next latent vector ($\hat{\Bell}_{t+1}$). 
Note that the predicted state ($\mathbf{s}_{t}^p$) is used by the decoder in late fusion not by the GRU.
However, in the early fusion architecture, the GRU has access to $\mathbf{s}_{t}^c,\ \mathbf{s}_{t}^p$ and $\Bell_{t}$ to predict $\hat{\Bell}_{t+1}$. 
The above results show two main points: first, the importance of adding model-based information to end-to-end learning architectures; and, second, the significance of where to inject the model-based information.

\subsection{Reinforcement Learning}

We followed the same architecture shown in Fig.~\ref{fig:action_network} for the reinforcement learning task.
The backbone networks (\gls{KARNet}) were frozen, and only the action network was trained using the temporal difference error from the DDPG algorithm.
Note that the action network is already pretrained using imitation learning, reducing the computational burden during reinforcement learning.
As shown in Table~\ref{tab:results_reward}, the early-fusion \gls{KARNet} performance is better than late-fusion \gls{KARNet}, reflecting the results of imitation learning.

\begin{table}[htbp]
	\renewcommand{\arraystretch}{1.5}
	\centering
	\caption{Reinforcement learning reward comparison.}
	\begin{tabular}{lcc}
		\hline
		\textbf{Model}  & \textbf{Early Fusion} & \textbf{Late Fusion} \\
		\hline
		\textbf{Reward} & 6324                  & 715.2                \\
		\hline
	\end{tabular}%
	\label{tab:results_reward}%
\end{table}%

\subsection{Udacity Dataset Evaluation}

The proposed \gls{KARNet} (early-fusion) showed a better performance margin on real-world data (Udacity dataset). 
In sim-to-real training, the network was pretrained on the CARLA dataset and then fine-tuned on the Udacity dataset.
Moreover, in sim-to-real training, when the network was pre-trained on a CARLA dataset and then fine-tuned on the Udacity dataset, the network consistently showed noticeably better performance compared to when trained entirely on real-world data.
When IMU and GNSS data is added to \gls{KARNet}, the performance improves significantly 
(Table~\ref{tab:results_imitation_latent_real}).

\begin{table}[htbp]
	\centering
	\begin{threeparttable}
		\renewcommand{\arraystretch}{1.5}
		\caption{Classification Performance using the real-world dataset (Udacity Dataset).}
		\begin{tabular}{p{1.5in}c}
			\hline
			\textbf{Model}                            & \textbf{Classification Accuracy\%} \\
			\hline
			\textbf{\gls{KARNet} without IMU \& GNSS} & 45.92 $\pm$ 1.66                   \\
			\textbf{\gls{KARNet} with IMU \& GNSS}    & \textbf{77.24} $\pm$ \textbf{1.24} \\
			\hline
		\end{tabular}%
		\label{tab:results_imitation_latent_real}%
	\end{threeparttable}
\end{table}%

Even though there is a significant gain in performance when IMU and GNSS are added, the over-all classification accuracy is not very high ($\approx77\%$).
A performance drop on the real dataset is expected since real data is more diverse, e.g., texture, shadows, camera over/under-exposure, etc. 
Additionally, since we explored the minimum possible configuration (i.e., the number of parameters and latent vector length), the autoencoder might lack the capacity to reconstruct real images compared to simulation data.

\section{Conclusion}

This paper presents a novel approach to learning latent dynamics in an autonomous driving scenario. 
The architecture combines end-to-end deep learning to learn the traffic dynamics from the front camera and model-based error-state extended Kalman filtering to model the ego vehicle.
The proposed architecture follows a dynamic convolutional autoencoder structure where the neural network consists of convolutional and recurrent architectures trained together. 
In this study, we investigate whether incorporating model-based information with model-free end-to-end learning can enhance performance, and how to integrate the two approaches effectively. 
To achieve this, we propose two methods: early-fusion, which involves incorporating state estimates obtained using a Kalman filter into the neural network architecture at an early stage; and late-fusion, which involves adding the vehicle state estimates 
at the end of the neural network architecture. 
We evaluated the effectiveness of our approach using both simulated and real-world driving data.
Our results indicate that integrating model-based information with model-free end-to-end learning significantly improves driving performance. Furthermore, we find that the early-fusion architecture is more effective than the late-fusion architecture. 
Moving forward, we plan to investigate the generalization and robustness properties of the proposed model in various more realistic and diverse scenarios, such as different towns or varying weather and lighting conditions.
Finally, most of our evaluations are simulator-based, using CARLA.
While simulator-based training provides a safe and controlled environment for testing and development, it is ultimately necessary to validate the system's performance on a real vehicle. 
For future studies, we plan explore the challenges and opportunities associated with transitioning (sim-to-real) from a simulator-based system to a real-world implementation for autonomous driving.

\bibliographystyle{IEEEtran}
\bibliography{references}

\end{document}